\documentclass[runningheads]{llncs}

\usepackage{eccv}

\usepackage{eccvabbrv}

\usepackage{graphicx}
\usepackage{svg}
\usepackage{booktabs}
\usepackage{multirow}
\usepackage{siunitx}
\usepackage{wrapfig}
\usepackage{adjustbox}
\usepackage{arydshln}
\usepackage{tabularx}
\usepackage{makecell}
\newcolumntype{Y}{>{\centering\arraybackslash}X}

\usepackage[accsupp]{axessibility}  

\usepackage{hyperref}

\usepackage{orcidlink}

\begin{document}

\title{MedPCFM: Improving Medical Point Cloud Completion by Integrating Point Transformers and Flow Matching} 

\titlerunning{MedPCFM: Medical Point Cloud Completion}

\author{Kamil Kwarciak\inst{1}\orcidlink{0000-0002-1392-4291} \and
Marek Wodzinski\inst{1,2}\orcidlink{0000-0002-8076-6246}}

\authorrunning{K.Kwarciak and M. Wodzinski}

\institute{Department of Measurement and Electronics, AGH University of Krakow, Krakow, Poland \and
Sano Centre for Computational Medicine, Krakow, Poland
\email{\{kwarciak,wodzinski\}@agh.edu.pl}}

\maketitle

\begin{abstract}
  Medical point cloud completion is important for anatomical reconstruction and downstream clinical workflows, yet generative modeling in this setting remains insufficiently studied. We investigate completion through continuous-time generative modeling and introduce PCFM, a PTv3-backed flow matching approach for medical point cloud completion. We evaluate on SkullFix and SkullBreak, and additionally on the more recent Mandibular Defect dataset. We build strong baselines by adapting PTv3 to a deterministic encoder-decoder completion model and by instantiating diffusion completion (PCDiff) with both PVCNN and PTv3 denoisers. PCFM with PTv3 is competitive with the deterministic PTv3 baseline and achieves state-of-the-art generative performance across datasets, while requiring substantially fewer sampling steps than diffusion. At the best operating points, PTv3 also yields clear throughput gains, providing up to a 7$\times$ speed-up for PCFM compared to a PVCNN backbone. Finally, we study empirical scaling trends by varying model size and point cardinality, showing consistent gains with higher point resolution and informative trade-offs across model scales.
  \keywords{Medical point cloud completion \and Flow matching \and 3D shape completion}
\end{abstract}

\begin{figure}[t]
  \centering
  \resizebox{\textwidth}{!}{\includegraphics{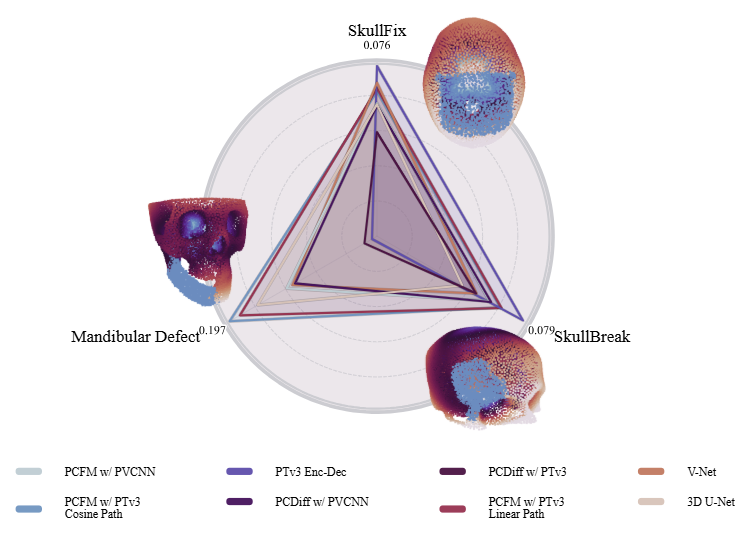}}
  \caption{Comparison of all methods evaluated in this study across all datasets using Chamfer Distance (CD) [mm], plotted on an inversely scaled radial axis so that lower CD indicates better performance while higher radial values improve visual interpretability.}
  \label{fig:visAbs}
\end{figure}

\section{Introduction}
Medical 3D reconstruction often requires completing partially observed anatomy, for example due to trauma, tumor resection, congenital defects, or limited scan coverage. While volumetric approaches have dominated many pipelines, point cloud representations are attractive for high-resolution geometry, efficient computation, and direct compatibility with surface reconstruction and downstream implant design. However, medical point cloud completion remains challenging: missing regions can be large, thin structures are common, and multiple anatomically consistent completions may exist for the same partial observation.

Recent progress in point cloud modeling has been driven by transformer backbones, with PointTransformerV3 (PTv3)~\cite{wu2024point} emerging as a state-of-the-art architecture through space-filling-curve serialization and efficient attention over serialized neighborhoods. In parallel, generative modeling has become a powerful tool for completion, with diffusion models demonstrating strong results for point sets in both natural~\cite{du2025superpc} and medical~\cite{friedrich2023point} domains. Yet diffusion-based sampling can be computationally expensive, and the scaling behavior of point cloud generative models in medical settings remains largely unexplored.

In this work, we study medical point cloud completion from a generative modeling perspective with an emphasis on flow matching. We develop an iterative pipeline that progressively strengthens both the backbone and the generative formulation: (i) we adapt PTv3 to a deterministic encoder-decoder completion setting and analyze reconstruction objectives; (ii) we study diffusion-based completion, reproducing state-of-the-art PVCNN-backed baseline~\cite{friedrich2023point} and replacing the denoiser with PTv3; and (iii) we introduce a PTv3-based flow matching model and ablate key design choices, including affine paths and an optional contrastive regularizer. We evaluate on the widely used SkullFix and SkullBreak benchmarks~\cite{kodym2021skullbreak}, and extend experiments to the recently introduced Mandibular Defect dataset~\cite{wu2025mandibular} to test generalizability. Beyond architecture and objectives, we further analyze practical trade-offs by ablating the number of steps in diffusion and flow matching, and we study an iterative completion strategy that aggregates predictions across multiple resampled inputs to boost point density for downstream surface reconstruction. Finally, we perform a systematic scaling study across model sizes and point cardinalities, providing empirical evidence of predictable performance trends with increased capacity and geometric resolution.

\textbf{Contributions.}
(i) We introduce a PTv3-based flow matching approach for medical point cloud completion and evaluate it on three medical defect datasets.
(ii) We adapt PTv3 to a deterministic encoder-decoder completion setting, establishing strong baselines and analyzing reconstruction objectives.
(iii) We provide a matched comparison between diffusion and flow matching for medical completion, including both PVCNN and PTv3 backbones, and analyze the impact of sampling-step budgets.
(iv) We report empirical scaling trends for medical point cloud generative modeling across model sizes and point cardinalities on SkullFix, SkullBreak, and Mandibular Defect.

\section{Related Works}

\subsection{Point Cloud Modeling}
Point cloud modeling studies learning on unordered sets of 3D points, aiming to extract geometric representations that are invariant to point permutations. Early deep learning models operated directly on points via shared MLPs and symmetric pooling (PointNet)~\cite{qi2017pointnet}, later extending to hierarchical local feature learning with neighborhood aggregation (PointNet++)~\cite{qi2017pointnet++} and attention-based point transformers, including Point Transformer and Point Transformer V2~\cite{zhao2021point,wu2022point}. More recent approaches adopt transformer backbones for stronger long-range reasoning on point sets, such as Point Transformer V3 (PTv3)~\cite{wu2024point}, which leverages space-filling-curve serialization and efficient attention to handle high-resolution point clouds. These models are widely used in 3D vision tasks such as classification~\cite{de2025interpretable,wu2025spiking}, semantic/instance segmentation~\cite{wang2018sgpn,vu2022softgroup}, and object detection~\cite{chen2023pimae,wang2022bridged}

\subsection{Medical Point Cloud Completion}
In medical imaging, point clouds naturally arise as surface samples extracted from segmented anatomy and provide a compact alternative to voxel grids while preserving fine geometric detail. Completion is relevant for reconstructive settings where anatomy is missing (e.g., trauma or resection) and surface models are used for surgical planning and patient-specific 3D-printed implant fabrication. Many learning-based pipelines have traditionally operated in voxel space using volumetric CNNs such as 3D U-Net and V-Net~\cite{cciccek20163d,milletari2016v}. More recently, medical defect completion has been studied with explicit surface/point representations, including adversarial point cloud reconstruction for cranial implant design (CranGAN)~\cite{sulakhe2022crangan}, supported by benchmarks such as AutoImplant challenge~\cite{li2023towards} and SkullFix/SkullBreak datasets~\cite{kodym2021skullbreak}. Recently, also novel generative approaches such as PCDiff~\cite{friedrich2023point} have been adapted to this field. Iterative point-cloud completion has also been explored to produce denser reconstructions that can benefit downstream meshing~\cite{wodzinski2023high}.

\subsection{Generative Models for Point Clouds}
Generative modeling provides a principled way to represent uncertainty and multi-modality in completion. Diffusion models (DDPMs)~\cite{ho2020denoising} have been adapted to point sets for unconditional generation and conditional completion~\cite{luo2021diffusion}, including point-voxel diffusion formulations such as PVD~\cite{zhou20213d}. In the medical implant setting, PCDiff applies diffusion to generate cranial implants from partial skull geometry and evaluates on SkullFix/SkullBreak~\cite{friedrich2023point}. In parallel, continuous-time alternatives such as Flow Matching~\cite{lipman2022flow} and related rectified-flow formulations~\cite{liu2022flow} train ODE-based generators by regressing velocity fields along probability paths, often enabling sampling with fewer solver steps than diffusion. While diffusion has seen broader adoption for point clouds~\cite{luo2021diffusion,zhou20213d,melas2023pc2}, flow-matching-style methods have only recently been explored in 3D point-cloud generation (e.g., GaussianAnything~\cite{yushi2025gaussiananything}) and remain comparatively under-explored for point set completion, motivating our investigation in the medical domain.

\section{Methods}
\begin{figure}[t]
  \centering
  \includegraphics[width=\linewidth]{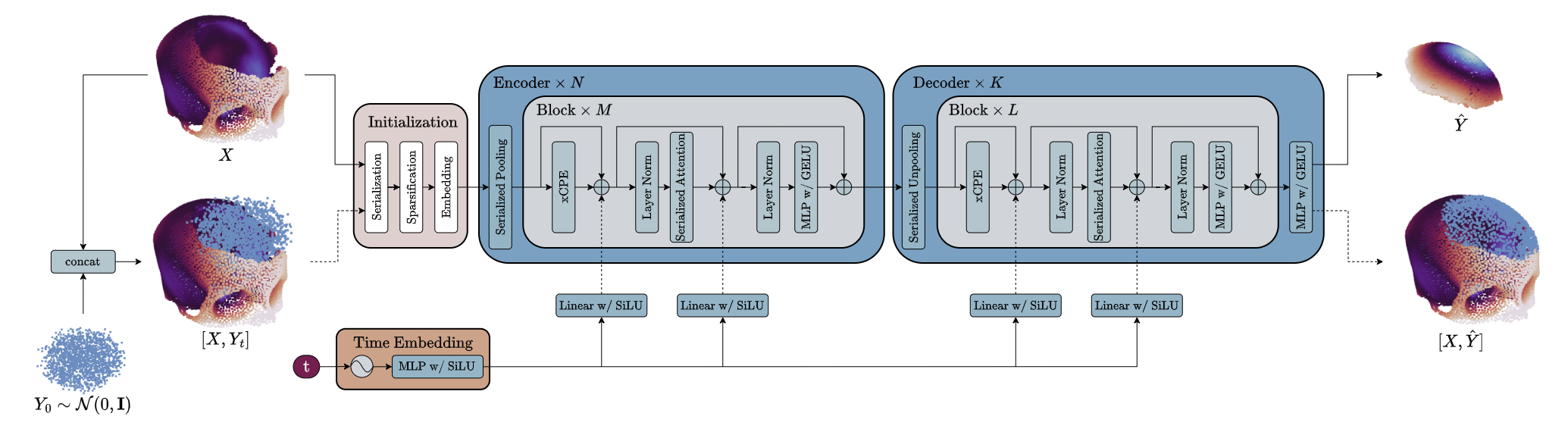}
  \caption{Overview of our pipeline. In the deterministic regime, the model maps the defective input $X$ to a completion $\hat{Y}$. In the generative regime, we model the missing part through a time-dependent state $Y_t$ conditioned on $X$: diffusion predicts $\epsilon_\theta(Y_t,t,X)$, while flow matching predicts a velocity field $v_\theta(Y_t,t,X)$ and generates $\hat{Y}$ by ODE integration. Dashed arrows indicate components used only in the generative regime.}
  \label{fig:architecture}
\end{figure}

We propose \textbf{PCFM}, a PTv3-based flow matching approach for medical point cloud completion. PCFM models the missing anatomy as a continuous-time transport from a simple noise distribution to the completion distribution conditioned on the observed defective input. We use Point Transformer V3 (PTv3)~\cite{wu2024point} as the backbone to parameterize the flow field over point sets. For completeness and fair comparison, we also introduce a deterministic PTv3 encoder-decoder baseline for medical completion and consider diffusion baselines following PCDiff~\cite{friedrich2023point} with both PVCNN and PTv3 denoisers. Additionally, we instantiate PCFM with a PVCNN backbone to isolate the impact of the backbone under the same flow-matching formulation. Figure~\ref{fig:architecture} summarizes the our proposed methodology.

\subsection{Problem Setup: Medical Point Cloud Completion}
Let $S \subset \mathbb{R}^3$ denote an anatomical surface and let $P=\{p_i\}_{i=1}^{N}\subset\mathbb{R}^3$ be a point cloud sampled from $S$. In completion, we observe an incomplete point cloud $X=\{x_i\}_{i=1}^{N_X}$ obtained by removing a subset of points (e.g., due to anatomical defects or partial acquisition), and aim to recover the missing geometry. We denote the missing part as $Y=\{y_j\}_{j=1}^{N_Y}$ such that the complete shape is $P=X\cup Y$ and $X\cap Y=\emptyset$. In practice, point clouds are treated as unordered sets, and this equality holds up to sampling density. Given only the observed points $X$, a model $f_\theta$ predicts the completion $\hat{Y}=f_\theta(X)$, yielding the reconstruction $\hat{P}=X\cup\hat{Y}$.

We can frame this problem in a deterministic regime, where we learn a direct mapping $f_\theta: X \mapsto \hat{Y}$ by minimizing a reconstruction loss $\mathcal{L}_{rec}(Y,\hat{Y})$. In the generative regime, we instead learn a conditional distribution over completions $p_{\theta}(Y\,|\,X)$ and sample $\hat{Y} \sim p_\theta(\cdot\,|\,X)$, which is relevant when the missing region is ambiguous and multiple anatomically consistent completions may exist.

Unless stated otherwise, we fix the output cardinality to $N_Y$ points. For the deterministic PTv3 encoder–decoder, we adopt an almost symmetric design and set $N_Y = N_X$
 to ensure comparable input and output representations and to simplify the optimization process. In the generative regime, we often choose $N_Y < N_X$ to reflect medical defects, where the missing anatomy typically occupies a smaller spatial extent, and to maintain a more balanced spatial density between $X$ and $\hat{Y}$.

\subsection{Backbone: Point Transformer V3 (PTv3)}
We use PTv3~\cite{wu2024point} as the shared point-cloud backbone across our point-based models. PTv3 addresses the unordered nature of point sets by serializing points with a space-filling curves, which induce a structured 1D ordering together with efficient neighborhood mapping. Self-attention is then performed as Serialized Attention, i.e., windowed/patch-wise attention over the serialized sequence, providing scalable local-to-global context aggregation without expensive neighbor recomputation. To inject geometric locality that complements attention, PTv3 employs xCPE (enhanced conditional positional encoding), implemented as a lightweight sparse-convolution module applied before attention. Finally, PTv3 uses Serialized Pooling and Serialized Unpooling, which down/up-sample points while preserving the serialization-based correspondence, enabling efficient multi-scale feature fusion with skip connections for high-resolution reasoning.

\subsection{Proposed Method: PTv3 Flow Matching for Completion (PCFM)}
We now describe our main method, PCFM, which learns a conditional continuous-time flow for the missing part given the observed input $X$.

\subsubsection{Flow Matching Formulation for Point Clouds}
\label{sec:fm_formulation}
We model the missing part as a point set $Y\in\mathbb{R}^{N_Y\times 3}$ and learn a conditional continuous-time flow that transforms a simple base distribution into the completion distribution given $X$. Let $Y_0 \sim \pi_0$ denote a base sample (we use an isotropic Gaussian in $\mathbb{R}^{N_Y\times 3}$) and let $Y_1 \sim \pi_1(\cdot\,|\,X)$ denote a ground-truth completion. Following a common FM parameterization, we define a time-dependent probability path via
\begin{equation}
Y_t \;=\; \psi_t(Y_0 \mid Y_1) \;=\; \alpha_t\,Y_1 \;+\; \sigma_t\,Y_0,
\end{equation}
where $\alpha_t,\sigma_t \in \mathbb{R}$ satisfy $\alpha_0=0,\sigma_0=1$ and $\alpha_1=1,\sigma_1=0$, so that $Y_{t=0}=Y_0$ and $Y_{t=1}=Y_1$. This induces a target velocity along the path
\begin{equation}
u_t \;=\; \frac{dY_t}{dt} \;=\; \dot{\alpha}_t\,Y_1 \;+\; \dot{\sigma}_t\,Y_0.
\end{equation}
Flow matching learns a conditional vector field $v_\theta(\cdot,t,X)$ such that the ODE
\begin{equation}
\frac{dY_t}{dt} \;=\; v_\theta(Y_t,t,X)
\end{equation}
transports $\pi_0$ to $\pi_1(\cdot\,|\,X)$. Training regresses the model velocity to the target velocity along the path~\cite{lipman2022flow}:
\begin{align}
\mathcal{L}_{FM} \;=\;
\mathbb{E}_{t\sim \mathcal{U}[0,1],\,Y_0\sim \pi_0,\,Y_1\sim \pi_1(\cdot|X)}
\Big[ \| v_\theta(Y_t,t,X) - u_t \|_2^2 \Big], \\
Y_t \;=\; \alpha_t Y_1 \;+\; \sigma_t Y_0.
\end{align}
At inference, we draw $Y_0\sim\pi_0$ and numerically integrate the learned ODE from $t=0$ to $t=1$ to obtain $\hat{Y}$, yielding $\hat{P}=X\cup\hat{Y}$. We parameterize $v_\theta$ with PTv3 to model the velocity field over point sets.

\subsubsection{Affine Paths for Point Cloud Flow Matching}
We introduced the affine path $Y_t=\alpha_t Y_1+\sigma_t Y_0$ and target velocity $u_t=\dot{\alpha}_t Y_1+\dot{\sigma}_t Y_0$. The remaining design choice is the scheduler $(\alpha_t,\sigma_t)$, which controls how signal and noise are mixed over time. While the linear path (Conditional-OT) with $\alpha_t = t$ and $\sigma_t = 1-t$ provides a baseline, we also consider alternative affine schedulers~\cite{lipman2024flow}. VP linear with the scheduler in the form of
\begin{equation}
\alpha_t = t,
\qquad
\sigma_t = \sqrt{1-t^2},
\end{equation}
and Cosine with the scheduler formed as
\begin{equation}
\alpha_t \;=\; \sin\!\Big(\tfrac{\pi}{2}t\Big),
\qquad
\sigma_t \;=\; \cos\!\Big(\tfrac{\pi}{2}t\Big).
\end{equation}
We report a detailed ablation of these path choices in the supplementary material, and use the two best-performing schedulers in the main experiments.

\subsubsection{Point Cloud Contrastive Flow Matching}
We optionally incorporate a contrastive regularizer inspired by $\Delta$FM~\cite{stoica2025contrastive}. For each example we form $Y_t=\alpha_t Y_1+\sigma_t Y_0$ and predict $v_\theta(Y_t,t,X)$. We then permute the batch to obtain a negative pair $(X^{-},Y_t^{-})$ and compute $v_\theta(Y_t^{-},t,X^{-})$. The objective is
\begin{equation}
\begin{aligned}
\mathcal{L}_{\Delta FM}
&=\mathbb{E}\Big[\|v_\theta(Y_t,t,X)-u_t\|_2^2\Big] \\
&\quad-\lambda\,\mathbb{E}\Big[\|v_\theta(Y_t,t,X)-v_\theta(Y_t^{-},t,X^{-})\|_2^2\Big].
\end{aligned}
\end{equation}
This term can encourage separation between mismatched conditional flows. We provide the corresponding ablation in the supplementary material, and keep the inference procedure unchanged (negatives are formed by within-batch permutation during training only).

\subsection{Baselines and Alternative Regimes}
We now summarize the deterministic and diffusion baselines used for comparison.

\subsubsection{Deterministic PTv3 Encoder-Decoder Baseline}
Given $X=\{x_i\}_{i=1}^{N_X}$, we embed points and process them with a hierarchical PTv3 encoder that serializes points and computes attention over serialized neighborhoods while progressively downsampling to multi-scale features. We attach a symmetric decoder with skip connections to upsample and fuse features, and a final prediction head regresses 3D coordinates to produce $\hat{Y}=\{\hat{y}_j\}_{j=1}^{N_Y}$, yielding $\hat{P}=X\cup\hat{Y}$.

We employ a two-term reconstruction loss. The first term is Chamfer Distance:
\begin{equation}
\mathcal{L}_{CD}(Y,\hat{Y})=
\frac{1}{|Y|}\sum_{y\in Y}\min_{\hat{y}\in \hat{Y}}\|y-\hat{y}\|_2^2
+
\frac{1}{|\hat{Y}|}\sum_{\hat{y}\in \hat{Y}}\min_{y\in Y}\|\hat{y}-y\|_2^2.
\end{equation}
To discourage point collapse and improve spatial coverage of the predicted completion, we add a repulsion (uniformity) regularizer following PU-Net~\cite{yu2018pu}:
\begin{equation}
\mathcal{L}_{rep}(\hat{Y})=
\frac{1}{|\mathcal{A}|}\sum_{\hat{y}_i\in \mathcal{A}}
\sum_{\hat{y}_j\in \mathcal{N}_k(\hat{y}_i)}
\Big(-r_{ij}\Big)\exp\!\Big(-\frac{r_{ij}^2}{h^2}\Big),
\qquad r_{ij}=\|\hat{y}_i-\hat{y}_j\|_2,
\end{equation}
where $\mathcal{A}\subseteq \hat{Y}$ is a set of $S$ anchors, $\mathcal{N}_k$ denotes $k$ nearest neighbors in $\hat{Y}$, and $h$ controls the repulsion radius. The final objective is
\begin{equation}
\label{eq:rec_loss}
\mathcal{L}_{rec}=\mathcal{L}_{CD}(Y,\hat{Y})+\lambda\,\mathcal{L}_{rep}(\hat{Y}),
\end{equation}
where $\lambda$ controls the repulsion strength.

\subsubsection{Diffusion Completion (PCDiff) and Backbone Variants}
We consider diffusion-based completion following PCDiff~\cite{friedrich2023point}, where the diffusion process acts on the missing part conditioned on $X$. Let $Y_1 = Y$ denote the clean missing point set and $Y_0\sim\mathcal{N}(0,\mathbf{I})$ a base noise sample in $\mathbb{R}^{N_Y\times 3}$. The forward process is
\begin{align}
q(Y_{1:T}\mid Y_1) &= \prod_{t=1}^{T} q(Y_t \mid Y_{t-1}), \\
q(Y_t \mid Y_{t-1}) &= \mathcal{N}\!\left(\sqrt{a_t}\,Y_{t-1},\, (1-a_t)\mathbf{I}\right),
\end{align}
with $a_t = 1-\beta_t$ and $\bar{a}_t=\prod_{s=1}^{t} a_s$. The marginal admits
\begin{equation}
Y_t \;=\; \sqrt{\bar{a}_t}\,Y_1 \;+\; \sqrt{1-\bar{a}_t}\,\epsilon,
\qquad \epsilon \sim \mathcal{N}(0,\mathbf{I}).
\end{equation}
To align with the FM affine notation, define
\begin{equation}
\alpha_t \;=\; \sqrt{\bar{a}_t},
\qquad
\sigma_t \;=\; \sqrt{1-\bar{a}_t},
\end{equation}
so that
\begin{equation}
Y_t \;=\; \alpha_t\,Y_1 \;+\; \sigma_t\,Y_0,
\qquad Y_0 \sim \mathcal{N}(0,\mathbf{I}).
\end{equation}
The reverse process is
\begin{equation}
p_\theta(Y_{0:T}\mid X) \;=\; p(Y_T)\prod_{t=T}^{1} p_\theta(Y_{t-1}\mid Y_t, X), 
\qquad p(Y_T)=\mathcal{N}(0,\mathbf{I}),
\end{equation}
with transitions parameterized by $\epsilon_\theta(Y_t, t, X)$. Using DDPM~\cite{ho2020denoising},
\begin{equation}
Y_{t-1}=\frac{1}{\sqrt{a_t}}
\left(
Y_t-\frac{1-a_t}{\sqrt{1-\bar{a}_t}}\epsilon_\theta(Y_t, t, X)
\right)
+\sqrt{\beta_t}\,z,
\qquad z\sim\mathcal{N}(0,\mathbf{I}),
\end{equation}
and training minimizes
\begin{equation}
\mathcal{L}_{diff} \;=\; 
\mathbb{E}_{t,\epsilon}\left[\left\|\epsilon - \epsilon_\theta(Y_t, t, X)\right\|_2^2\right].
\end{equation}

We instantiate $\epsilon_\theta$ with (i) the PVCNN denoiser used in PCDiff (PointNet++ with point-voxel convolutions)~\cite{qi2017pointnet++,liu2019point}, and (ii) a PTv3 denoiser while keeping the diffusion parameterization, schedule, and sampling unchanged.

\section{Experiments}
We conduct experiments to (i) compare deterministic and generative completion regimes, (ii) quantify the impact of backbone and loss design, and (iii) study scaling with model capacity and point cardinality. Details are organized in the following subsections.

\subsection{Datasets}
Medical datasets for defect-driven shape completion are still emerging. We focus on two widely used cranial defect benchmarks, SkullFix and SkullBreak~\cite{kodym2021skullbreak}, and additionally evaluate on the recently introduced Mandibular Defect dataset~\cite{wu2025mandibular}, which contains 147 clinically derived mandibular defect cases. For SkullFix and SkullBreak, we use the official train/test splits and further split the training set into train/validation, resulting in 90/10/110 (train/val/test) for SkullFix and 510/60/100 for SkullBreak. Given the dataset size disparity, we follow a joint training regime~\cite{kwarciak2023deep} across SkullFix and SkullBreak to encourage better cross-dataset generalization. This differs from PCDiff~\cite{friedrich2023point}, where SkullFix and SkullBreak were trained separately. For Mandibular Defect dataset, no official split is provided, hence we adopt a 70/10/20 split, yielding 103/14/30 cases.

We convert surface representations to dense point clouds via Poisson-based surface sampling and obtain the target training cardinalities using farthest point sampling. We evaluate three point cardinalities. For the deterministic PTv3 encoder-decoder, we use $(N_X/N_Y)\in\{2048/2048,\;16384/16384,\;32768/32768\}$. For generative models, we predict fewer points for the missing region to reflect the typically smaller spatial extent of defects and maintain comparable point density, using $(N_X/N_Y)\in\{1844/204,\;14746/1638,\;29492/3276\}$.

Following PCDiff~\cite{friedrich2023point}, we normalize all shapes by applying a global scaling to map coordinates into the range $[-3,3]$ along each axis. During training, we apply a lightweight rigid augmentation consisting of fixed rotations around the coordinate axes to improve invariance to global pose.

\subsection{Experimental Setup}
We begin with the deterministic PTv3 encoder-decoder baseline and study three model scales (small/base/large) with 19.7M, 46.7M, and 468.6M parameters, respectively. Full architectural specifications are provided in the supplementary material for reproducibility. For reconstruction, we train with Chamfer Distance and additionally ablate a repulsion regularizer (Eq.~\ref{eq:rec_loss}) using two strengths, setting $\lambda\in\{30,300\}$. Initial experiments for the encoder-decoder are performed at the 2048/2048 and 16384/16384 cardinalities. We train for 10,000 epochs with ADAM optimizer~\cite{kingma2014adam} and adjust batch size to the cardinality (32 for 2048/2048 and 8 for 16384/16384), monitoring training/validation loss to ensure convergence.

For generative completion, we start from the PCDiff baseline with a PVCNN denoiser~\cite{friedrich2023point} (27.6M parameters) and follow the training protocol of the original work. We then isolate backbone effects by replacing PVCNN with PTv3-base (46.7M) while keeping the diffusion formulation and schedule fixed. Next, we move to flow matching and, for fair comparison, train a conditional-OT FM baseline with a PVCNN backbone. All flow matching ablations (affine path choices and contrastive regularization) are conducted with PTv3. To study scaling, we evaluate all three point cardinalities and all three model sizes. Diffusion and flow matching models are trained for 15,000 epochs (as in PCDiff~\cite{friedrich2023point}), with ADAM optimizer~\cite{kingma2014adam} and with minor backbone-dependent adjustments to batch size and learning rate reported in the supplementary material.

All experiments are run on a single NVIDIA GH200 GPU with 96GB VRAM.

\subsection{Evaluation Protocol, Metrics, and Baselines}
We evaluate our approach in three stages. First, we validate the deterministic PTv3 encoder-decoder baseline by analyzing the effect of reconstruction objectives across model sizes and point cardinalities. Second, we study the generative regime, focusing on flow matching: we ablate affine paths and the contrastive component, and quantify scaling trends with respect to model size and point cardinality using Chamfer Distance (CD). Third, we compare the best-performing flow matching model against diffusion baselines on all datasets. For SkullFix/SkullBreak we train jointly and evaluate on each benchmark separately, while for Mandibular Defect we train and evaluate only within that dataset.

\begin{wraptable}[23]{r}{0.52\textwidth}
\centering
\scriptsize
\setlength{\tabcolsep}{2.0pt}
\renewcommand{\arraystretch}{1.}
\caption{PTv3 encoder-decoder ablation across model size, point cardinality ($N_X/N_Y$), and repulsion strength $\lambda$. Lower CD [mm] is better.}
\begin{tabular}{llr
                S[table-format=1.4]
                S[table-format=1.4]}
\toprule
& & &
\multicolumn{1}{c}{\textbf{SkullFix}} &
\multicolumn{1}{c}{\textbf{SkullBreak}} \\
\cmidrule(lr){4-4}\cmidrule(lr){5-5}
\textbf{Model} & $\boldsymbol{N_X/N_Y}$ & $\boldsymbol{\lambda}$ &
\multicolumn{1}{c}{CD$\downarrow$} &
\multicolumn{1}{c}{CD$\downarrow$} \\
\midrule

\multirow{6}{*}{Small}
& \multirow{3}{*}{2048/2048}   & 0   & 0.125 & 0.132 \\
&                              & 30  & 0.123 & 0.134 \\
&                              & 300 & 0.135 & 0.147 \\
\cmidrule(lr){2-5}
& \multirow{3}{*}{16384/16384} & 0   & 0.110 & 0.108 \\
&                              & 30  & 0.114 & 0.111 \\
&                              & 300 & 0.130 & 0.109 \\
\midrule

\multirow{6}{*}{Base}
& \multirow{3}{*}{2048/2048}   & 0   & 0.105 & 0.116 \\
&                              & 30  & 0.111 & 0.117 \\
&                              & 300 & 0.115 & 0.131 \\
\cmidrule(lr){2-5}
& \multirow{3}{*}{16384/16384} & 0   & \textbf{0.076} & 0.079 \\
&                              & 30  & 0.077 & 0.083 \\
&                              & 300 & 0.078 & \textbf{0.078} \\
\midrule

\multirow{6}{*}{Large}
& \multirow{3}{*}{2048/2048}   & 0   & 0.118 & 0.123 \\
&                              & 30  & 0.116 & 0.116 \\
&                              & 300 & 0.139 & 0.137 \\
\cmidrule(lr){2-5}
& \multirow{3}{*}{16384/16384} & 0   & 0.089 & 0.087 \\
&                              & 30  & 0.095 & 0.086 \\
&                              & 300 & 0.103 & 0.093 \\
\bottomrule
\end{tabular}
\label{tab:ptv3_ed}
\end{wraptable}

We report metrics across point-cloud and volumetric 3D representations. For point clouds, we use Chamfer Distance (CD) as the primary metric. For comparison to volumetric baselines, we convert our point-based reconstructions to voxel grids by first reconstructing a watertight surface/occupancy using Shape-As-Points (SAP), an optimization-based differentiable Poisson solver~\cite{peng2021shape} and then voxelizing the resulting representation at the evaluation resolution. We then report DSC, boundary DSC (BDSC), and HD95, alongside 3D U-Net~\cite{cciccek20163d} and V-Net~\cite{milletari2016v} baselines.

\subsection{Results}

\begin{figure}[t]
  \centering
  \includegraphics[width=0.92\linewidth]{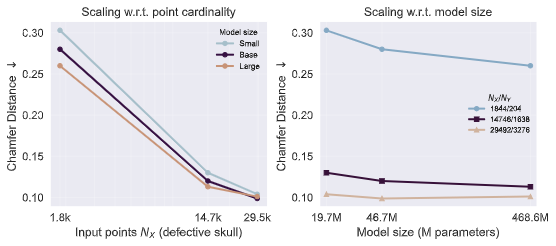}
  \caption{Scaling trends for PCFM on SkullBreak. Left: Chamfer Distance [mm] versus input point cardinality $N_X$. Right: Chamfer Distance [mm] versus model size (parameter count). Lower is better.}
  \label{fig:scaling}
\end{figure}

\begin{table}[t]
\centering
\small
\setlength{\tabcolsep}{3.5pt}
\renewcommand{\arraystretch}{1.20}
\caption{Main comparison across datasets and representations. Lower is better for CD [mm] and HD95 [mm]; higher is better for DSC and BDSC. V-Net and 3D U-Net are volumetric baselines operating on voxel grids.}
\label{tab:gen_main}

\resizebox{\textwidth}{!}{%
\begin{tabular}{lcccc cccc cccc}
\toprule
& \multicolumn{4}{c}{\textbf{SkullFix}}
& \multicolumn{4}{c}{\textbf{SkullBreak}}
& \multicolumn{4}{c}{\textbf{Mandibular Defect}} \\
\cmidrule(lr){2-5}\cmidrule(lr){6-9}\cmidrule(lr){10-13}
\textbf{Model}
& CD$\downarrow$ & DSC$\uparrow$ & BDSC$\uparrow$ & HD95$\downarrow$
& CD$\downarrow$ & DSC$\uparrow$ & BDSC$\uparrow$ & HD95$\downarrow$
& CD$\downarrow$ & DSC$\uparrow$ & BDSC$\uparrow$ & HD95$\downarrow$ \\
\midrule

PCDiff w/ PVCNN              & 0.141 & 0.821 & 0.801 & 2.83  & 0.140 & 0.818 & 0.805 & 2.46  & 0.267 & 0.571 & 0.578 & 5.81 \\
PCDiff w/ PTv3               & 0.186 & 0.744 & 0.721 & 4.58  & 0.169 & 0.759 & 0.720 & 3.34 & 0.341 & 0.461 & 0.441 & 8.31 \\

PCFM w/ PVCNN                & 0.138 & 0.823 & 0.759 & 3.04  & 0.143 & 0.821 & 0.778 & 2.88  & 0.258 & 0.596 & 0.590 & 5.85 \\
PCFM w/ PTv3 (Linear)        & 0.113 & 0.842 & 0.816 & 2.40  & 0.120 & \textbf{0.833} & 0.819 & \textbf{2.38}  & 0.208 & \textbf{0.713} & \textbf{0.721} & 3.28 \\
PCFM w/ PTv3 (Cosine)        & 0.118 & 0.835 & 0.807 & 2.53  & 0.122 & 0.826 & 0.802 & 2.43  & \textbf{0.197} & 0.639 & 0.659 & \textbf{3.05} \\

PTv3 Encoder-Decoder         & \textbf{0.076} & 0.854 & 0.841 & 2.33  & \textbf{0.079} & 0.814 & 0.813 & 2.51  & 0.349 & 0.434 & 0.444 & 9.21 \\
\hdashline
V-Net                        & 0.104 & \textbf{0.856} & 0.851 & \textbf{2.15}  & 0.168 & 0.822 & \textbf{0.837} & 2.48  & 0.264 & 0.546 & 0.550 & 7.11 \\
3D U-Net                     & 0.140 & 0.851 & \textbf{0.852} & 2.32  & 0.195 & 0.829 & 0.821 & 2.49  & 0.227 & 0.642 & 0.640 & 4.98 \\

\bottomrule
\end{tabular}%
}
\end{table}

We begin with the deterministic PTv3 encoder-decoder, which to our knowledge is the first adaptation of PTv3 to medical point cloud completion, and ablate reconstruction losses across model scales and point cardinalities (Table~\ref{tab:ptv3_ed}). Increasing the point count yields the largest gains, while scaling beyond PTv3-Base does not consistently improve performance. Additionally, overly strong repulsion can degrade accuracy (see supplementary for qualitative examples).

We then move to the generative regime and report a comprehensive comparison in Table~\ref{tab:gen_main}, including the PTv3 encoder-decoder, diffusion baselines (PCDiff with PVCNN and PTv3 backbones), flow matching baselines (PCFM with PVCNN), our two best PCFM with PTv3 variants, and volumetric references (V-Net and 3D U-Net). The selection of the top-performing PCFM with PTv3 configurations is based on ablations over affine paths and the contrastive component (reported in the supplementary), while Figure~\ref{fig:scaling} summarizes scaling trends in the generative regime with respect to model size and point cardinality. Overall, performance improves consistently with increasing the amount of points, whereas scaling model size exhibits diminishing returns beyond the base configuration, suggesting that resolution is the dominant driver in this setting. Table~\ref{tab:gen_main} compares point- and voxel-space metrics across datasets. On SkullFix and SkullBreak, the deterministic PTv3 encoder-decoder achieves the lowest CD, which is expected since Chamfer is its direct training objective, while PCFM with PTv3 provides the strongest generative performance and remains competitive on DSC/BDSC/HD95. On Mandibular Defect, PCFM with PTv3 yields the best overall results among the compared methods, while volumetric baselines underperform; we attribute this partly to the dataset’s higher geometric complexity and the practical difficulty of training volumetric models at high resolution. Since optimizing volumetric architectures is not the primary focus of this work, our voxel baselines follow standard settings and limited tuning, and we expect that more extensive volumetric optimization could further improve their performance.

\begin{wraptable}[24]{r}{0.52\textwidth}
\centering
\scriptsize
\setlength{\tabcolsep}{2.0pt}
\renewcommand{\arraystretch}{1.}
\caption{Sampling-step trade-offs for diffusion (PCDiff uses the DDPM ancestral sampler) and flow matching (PCFM uses a Heun ODE solver) on the combined SkullFix/SkullBreak setting. Time is reported per generated sample, lower CD [mm] is better.}
\label{tab:sampling_steps}
\begin{tabular}{ll S[table-format=4.0] S[table-format=3.3] S[table-format=1.3]}
\toprule
\textbf{Method} & \textbf{Backbone} & {\textbf{Steps}} & {\textbf{Time [s]}} & {\textbf{CD} $\downarrow$} \\
\midrule
\multirow{8}{*}{PCDiff}
& \multirow{4}{*}{PVCNN}
& 1    & 0.550   & 2.440 \\
&      & 50   & 27.410  & 1.970 \\
&      & 100  & 55.040  & 1.770 \\
&      & 1000 & 553.860 & \textbf{0.141} \\
\cmidrule(lr){2-5}
& \multirow{4}{*}{PTv3}
& 1    & 0.276  & 2.430 \\
&      & 50   & 6.700  & 2.270 \\
&      & 100  & 9.270  & 2.150 \\
&      & 1000 & 99.800 & \textbf{0.178} \\
\midrule
\multirow{8}{*}{PCFM}
& \multirow{4}{*}{PVCNN}
& 1  & 1.090  & 0.486 \\
&    & 5  & 4.400  & 0.180 \\
&    & 20 & 20.770 & 0.142 \\
&    & 40 & 42.800 & \textbf{0.140} \\
\cmidrule(lr){2-5}
& \multirow{4}{*}{PTv3}
& 1  & 0.157 & 0.478 \\
&    & 5  & 0.643 & 0.166 \\
&    & 20 & 2.770 & 0.120 \\
&    & 40 & 6.100 & \textbf{0.116} \\
\bottomrule
\end{tabular}
\end{wraptable}

Sampling efficiency is a central practical difference between diffusion and flow matching. Flow matching learns a continuous-time ODE whose solution can often be approximated with far fewer integration steps than the number of denoising steps typically required for diffusion sampling. For diffusion, we follow PCDiff and use the DDPM ancestral sampler. For flow matching, we integrate the ODE with Heun’s method, which has been reported as a robust choice in recent vision work~\cite{ma2024sit}. We evaluate diffusion with 1, 50, 100, and 1000 sampling steps, and flow matching with 1, 5, 20, and 40 integration steps. Since one-step generation remains an active research direction for both diffusion and flow-based methods~\cite{noroozi2024you,you2025consistency,kornilov2024optimal} and often requires specialized training, we treat it as a diagnostic setting rather than the primary operating regime. Table~\ref{tab:sampling_steps} reports the resulting speed-quality trade-offs. At the best-performing settings for each method (PCDiff at 1000 steps and PCFM at 40 Heun steps), PCFM with PTv3 achieves the most favorable combination of speed and accuracy. Relative to the corresponding PVCNN variants, PTv3 provides a $\sim 5.5\times$ speed-up for PCDiff (1000 steps) and a $\sim 7\times$ speed-up for PCFM (40 steps), while maintaining strong reconstruction quality. Overall, diffusion requires long sampling to reach competitive CD, whereas flow matching attains comparable or better accuracy with substantially fewer steps, making PCFM with PTv3 the most practical choice in our setting.

\subsection{Qualitative Analysis}
We complement quantitative results with qualitative comparisons of the generated point clouds (Figure~\ref{fig:qualitative}, additional examples are provided in the supplementary). For the volumetric baselines (V-Net and 3D U-Net), we extract surfaces and sample point clouds using the same pipeline as for our point-based methods (Poisson surface sampling followed by farthest point sampling). Overall, all methods produce visually plausible completions on SkullFix and SkullBreak, with differences largely reflecting point density and sampling. In particular, the PTv3 encoder-decoder outputs denser implants due to the $N_X=N_Y$ setting. We note that the visually uniform point distribution for volumetric baselines is partly induced by the post-processing (FPS on Poisson-sampled surfaces) rather than being directly predicted as points.

On Mandibular Defect, qualitative differences are more pronounced: PCFM with PTv3 most closely matches the ground-truth geometry, while PCDiff with PVCNN can reach similar fidelity but at the cost of long sampling (e.g., 1000 steps). Other methods typically recover the correct region but exhibit increased geometric noise in the predicted implant. We additionally observe more frequent artifacts for volumetric models on Mandibular Defect, which contains high-resolution scans (up to $\sim$1000$^3$ voxels). Training on downsampled volumes and converting predictions back to high resolution can lead to small spatial shifts and boundary jitter, especially given the limited dataset size.

\begin{figure}[h]
    \centering
    \includegraphics[width=\textwidth]{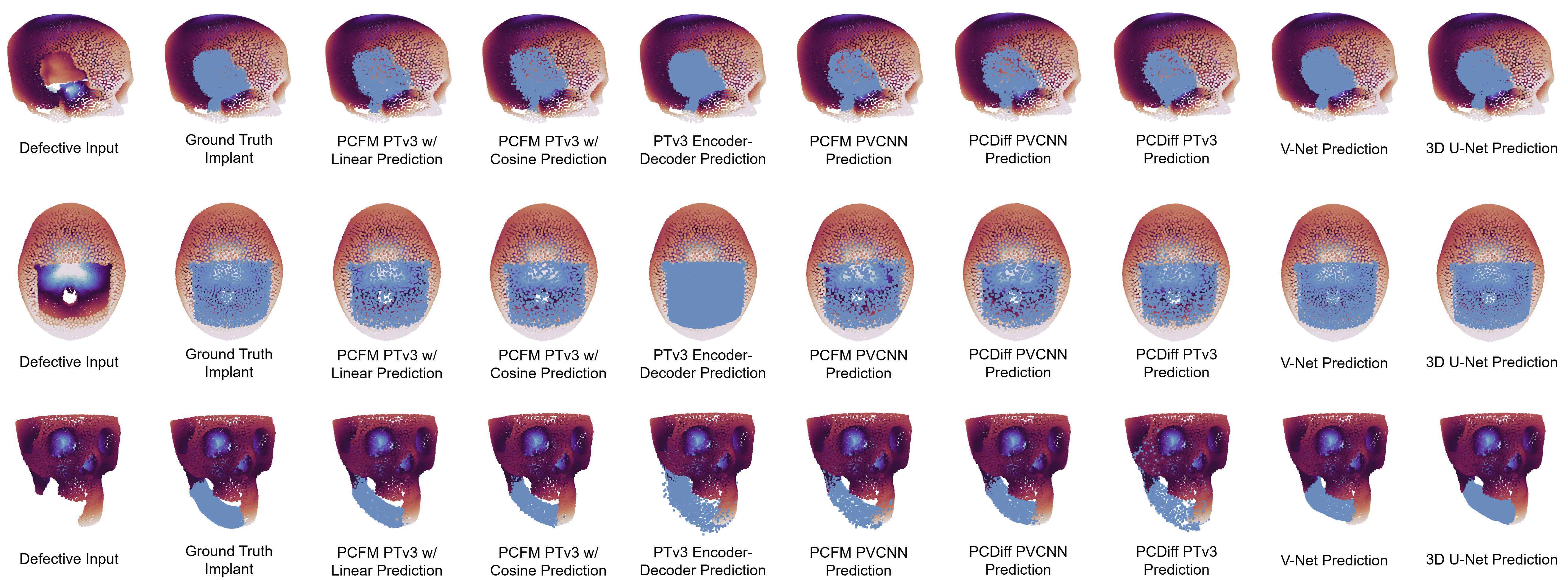}
    \caption{Qualitative comparison across datasets. Top: SkullBreak, middle: SkullFix, bottom: Mandibular Defect. For the volumetric baselines, point clouds are obtained by extracting surfaces and sampling points using Poisson surface sampling followed by farthest point sampling (FPS), consistent with the preprocessing used for point-based methods.}
    \label{fig:qualitative}
\end{figure}

\section{Discussion}
We introduced \textbf{PCFM} with a PTv3 backbone as a flow-matching approach to medical point cloud completion. We evaluated on two established cranial defect datasets (SkullFix and SkullBreak) and additionally on the Mandibular Defect dataset. Across these datasets, PCFM with PTv3 is the strongest generative approach in our study and remains competitive with volumetric models and robust deterministic PTv3 encoder-decoder baseline, which itself is a contribution as an adaptation of PTv3 to medical point cloud completion.

A key motivation for generative modeling in this setting is that implant generation is often not uniquely determined by the observed defect. Generative methods can provide multiple anatomically consistent candidates that can be selected or refined according to clinical constraints. In this context, the practical sampling efficiency of PCFM with PTv3 is particularly attractive, enabling rapid generation (e.g., a single sample in $\sim$6.1\,s in our setup) while maintaining strong geometric fidelity.

Our results also support point-based modeling as an efficient alternative to volumetric pipelines. Training and inference with dense 3D convolutions are compute- and memory-intensive, and voxelization introduces resolution trade-offs that can be limiting for fine anatomical detail. Point clouds provide a compact representation that is sufficient for downstream clinical workflows: implants are typically manufactured from surface meshes, and point predictions can be converted to meshes directly without requiring a voxel intermediate.

We also highlight limitations and future work. A practical bottleneck remains the conversion from point clouds to meshes (and subsequently to voxel grids for volumetric metrics). We use an optimization-based differentiable Poisson reconstruction, but mesh quality can be sensitive to reconstruction hyperparameters and point density. Future work could explore learning-based surface reconstruction~\cite{peng2021shape,friedrich2023point}, more robust meshing pipelines, or iterative multi-resampling completion (investigated as an ablation in the supplementary) to further stabilize mesh generation. In addition, Mandibular Defect yields lower point- and voxel-space scores compared to SkullFix/SkullBreak, likely reflecting its higher anatomical variability and geometric complexity. Notably, the deterministic PTv3 encoder-decoder degrades more strongly on Mandibular Defect, whereas PCFM with PTv3 provides a marked improvement; we hypothesize this is driven by the substantially higher heterogeneity of mandibular defects and the limited ability of non-generative mappings to generalize across such variability. Improving stability and generalization in such settings is an important direction for future works.

Overall, PCFM with PTv3 provides a simple and effective generative framework for medical point cloud completion, combining strong accuracy with fast sampling. We expect this formulation to be a useful foundation for future work on richer conditioning, improved surface reconstruction, and alternative continuous-time generative objectives for medical 3D shape reconstruction.

\section*{Acknowledgements}
The project was funded by The National Centre for Research and Development, Poland under Lider Grant no: LIDER13/0038/2022 (DeepImplant). We gratefully acknowledge Polish high-performance computing infrastructure PLGrid (HPC Center: ACK Cyfronet AGH) for providing computer facilities and support within computational grant no. PLG/2026/019392.

\bibliographystyle{splncs04}
\bibliography{main}

\clearpage
\appendix

\section{Architectures and Training Details}
Table~\ref{tab:ptv3_configs} reports the PTv3 architectural configurations used in this work together with the associated training hyperparameters for both deterministic and generative regimes. Per-stage lists are ordered from shallow to deep (i.e., from early high-resolution stages to later low-resolution stages), and we keep this ordering consistent across encoder and decoder specifications. Unless stated otherwise, these settings are used across all experiments to facilitate reproducibility.

\begin{table*}[h]
\centering
\footnotesize
\setlength{\tabcolsep}{4pt}
\renewcommand{\arraystretch}{1.12}
\caption{PTv3 architectural configurations and training hyperparameters for the Small, Base, and Large variants.}
\label{tab:ptv3_configs}
\begin{tabularx}{\textwidth}{@{}lYYY@{}}
\toprule
\textbf{Setting} & \textbf{Small} & \textbf{Base} & \textbf{Large} \\
\midrule

\multicolumn{4}{@{}l}{\textbf{Architecture}} \\
\cmidrule(l){1-4}
Space-filling curves
& \makecell[c]{Z, Z-trans}
& \makecell[c]{Z, Z-trans,\\ Hilbert,\\ Hilbert-trans}
& \makecell[c]{Z, Z-trans,\\ Hilbert,\\ Hilbert-trans} \\
Stride
& \makecell[c]{2, 2, 2, 2}
& \makecell[c]{2, 2, 2, 2}
& \makecell[c]{2, 2, 2, 2} \\
Encoder depths
& \makecell[c]{2, 2, 2, 4, 2}
& \makecell[c]{2, 2, 2, 6, 2}
& \makecell[c]{6, 6, 6, 10, 6} \\
Encoder channels
& \makecell[c]{24, 48,\\ 96, 192, 320}
& \makecell[c]{32, 64,\\ 128, 256, 512}
& \makecell[c]{64, 128,\\ 256, 512, 1024} \\
Encoder heads
& \makecell[c]{2, 2, 4, 8, 10}
& \makecell[c]{2, 4, 8, 16, 32}
& \makecell[c]{4, 8, 16, 32, 32} \\
Encoder patch size
& \makecell[c]{768, 768,\\ 768, 768, 768}
& \makecell[c]{1024, 1024,\\ 1024, 1024, 1024}
& \makecell[c]{2048, 2048,\\ 2048, 2048, 2048} \\
Decoder depths
& \makecell[c]{2, 2, 2, 2}
& \makecell[c]{2, 2, 2, 2}
& \makecell[c]{6, 6, 6, 6} \\
Decoder channels
& \makecell[c]{48, 48, 96, 192}
& \makecell[c]{64, 64, 128, 256}
& \makecell[c]{128, 128, 256, 512} \\
Decoder heads
& \makecell[c]{2, 2, 4, 8}
& \makecell[c]{4, 4, 8, 16}
& \makecell[c]{8, 8, 16, 16} \\
Decoder patch size
& \makecell[c]{768, 768,\\ 768, 768}
& \makecell[c]{1024, 1024,\\ 1024, 1024}
& \makecell[c]{2048, 2048,\\ 2048, 2048} \\
Stochastic depth
& 0.1 & 0.1 & 0.1 \\
\makecell[l]{Completion head\\ hidden channels}
& 384 & 512 & 1024 \\

\midrule
\multicolumn{4}{@{}l}{\textbf{Deterministic training}} \\
\cmidrule(l){1-4}
Learning rate
& $3 \times 10^{-4}$
& $3 \times 10^{-4}$
& $1 \times 10^{-4}$ \\
Weight decay
& $1 \times 10^{-4}$
& $1 \times 10^{-4}$
& $1 \times 10^{-4}$ \\
Epochs
& 10\,000 & 10\,000 & 10\,000 \\
Batch size
& 32 & 16 & 8 \\
\makecell[l]{Repulsion warmup\\ epochs}
& 200 & 200 & 200 \\

\midrule
\multicolumn{4}{@{}l}{\textbf{Generative training}} \\
\cmidrule(l){1-4}
Learning rate
& $1 \times 10^{-4}$
& $1 \times 10^{-4}$
& $1 \times 10^{-4}$ \\
Weight decay
& $1 \times 10^{-4}$
& $1 \times 10^{-4}$
& $1 \times 10^{-4}$ \\
Epochs
& 15\,000 & 15\,000 & 15\,000 \\
Batch size
& 32 & 16 & 8 \\
Warmup steps
& 2000 & 2000 & 2000 \\
EMA decay
& 0.999 & 0.999 & 0.999 \\
\bottomrule
\end{tabularx}
\end{table*}

\section{Statistical Significance Analysis for Repulsion Term in PTv3 Encoder-Decoder}
We evaluate whether the repulsion term significantly affects Chamfer Distance (CD) for the deterministic PTv3 encoder-decoder. For each dataset and each pair of repulsion strengths, we test the null hypothesis $H_0$ that the median of the paired per-case differences in CD is zero (i.e., repulsion does not change CD). Because CDs are computed on the same test cases for all settings, we use paired two-sided Wilcoxon signed-rank tests and apply Holm correction across the three pairwise comparisons per dataset. We consider results statistically significant at $\alpha=0.05$ after Holm correction (i.e., $p_{\mathrm{Holm}}<0.05$). To aid interpretation, we note that a significant result indicates a systematic shift in CD across cases rather than a change driven by a small number of outliers.

Table~\ref{tab:repulsion_stats} shows that on SkullBreak, repulsion does not yield statistically significant differences in CD under this setting. On SkullFix, $\lambda=30$ differs significantly from $\lambda=0$, and $\lambda=300$ differs significantly from $\lambda=30$ after Holm correction, while $\lambda=300$ is not significantly different from $\lambda=0$. Overall, the effect of repulsion appears dataset-dependent, supporting its treatment as an ablation rather than a default component.

\clearpage

\begin{table}[h]
\centering
\setlength{\tabcolsep}{10pt}
\caption{Paired Wilcoxon tests for repulsion strength $\lambda$ (Base 16384/16384). Holm-corrected $p$-values are reported per dataset.}
\label{tab:repulsion_stats}
\begin{tabular}{l l c c c}
\toprule
Dataset & Comparison & $n$ & $p$ & $p_{\mathrm{Holm}}$ \\
\midrule
SkullBreak & $0$ vs $30$   & 100 & 0.0901  & 0.2702 \\
SkullBreak & $0$ vs $300$  & 100 & 0.5247  & 0.5247 \\
SkullBreak & $30$ vs $300$ & 100 & 0.2369  & 0.4738 \\
\addlinespace[2pt]
SkullFix   & $0$ vs $30$   & 110 & 0.00813 & 0.0163 \\
SkullFix   & $0$ vs $300$  & 110 & 0.3498  & 0.3498 \\
SkullFix   & $30$ vs $300$ & 110 & 0.00127 & 0.00381 \\
\bottomrule
\end{tabular}
\end{table}

\section{Qualitative Effect of Repulsion Loss}
\begin{figure}[h]
  \centering
  \includegraphics[width=\linewidth]{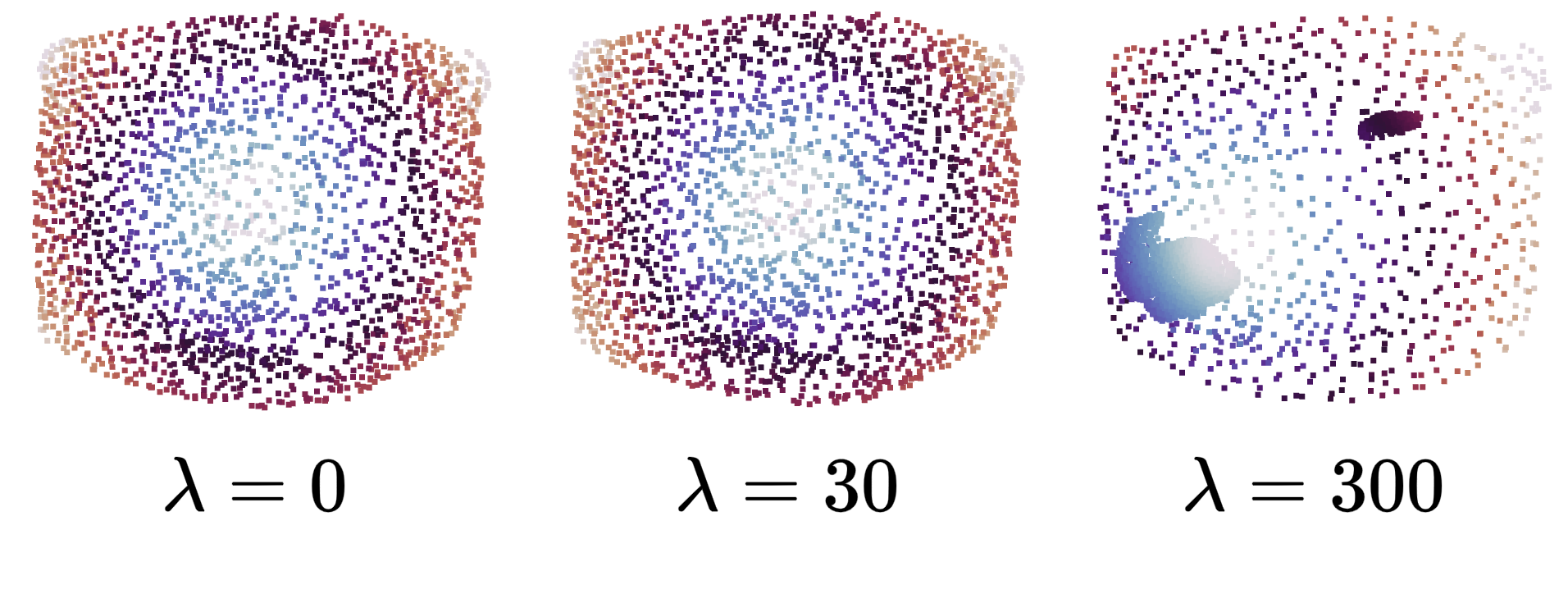}
  \caption{Qualitative effect of repulsion in the deterministic PTv3 encoder-decoder SkullFix (Small, 2048/2048). Mild repulsion ($\lambda=30$) yields a visually similar completion to $\lambda=0$, whereas strong repulsion ($\lambda=300$) can produce degenerate point distributions with large voids and concentrated point islands.}
  \label{fig:repulsion_vis}
\end{figure}

We further inspect the effect of repulsion qualitatively on SkullFix. Figure~\ref{fig:repulsion_vis} visualizes predictions for the Small model at the 2048/2048 setting for three repulsion strengths. The outputs for $\lambda=0$ and $\lambda=30$ are visually similar, suggesting limited qualitative impact of mild repulsion in this example. In contrast, a strong repulsion term ($\lambda=300$) leads to an undesirable failure mode: the predicted implant.
contains large low-density regions together with concentrated point islands, indicating that overly strong uniformity pressure can destabilize the reconstruction. We therefore treat repulsion as an ablation and use $\lambda=0$ as the default.

\clearpage

\section{Multi-Sample Generative Variability}
A practical advantage of generative completion is that the model can be sampled multiple times for the same defective input $X$ by drawing different initial noise samples. This yields a set of plausible implant predictions
\begin{equation}
\{\hat{Y}^{(k)}\}_{k=1}^{K}, 
\qquad 
\hat{Y}^{(k)} \sim p_{\theta}(\cdot \mid X),
\end{equation}
rather than a single deterministic output. Such multi-sample generation is useful in medical reconstruction because defect completion can be ambiguous, especially near defect boundaries, where multiple anatomically reasonable implant shapes may exist.

\begin{figure}[t]
  \centering
  \includegraphics[width=\linewidth]{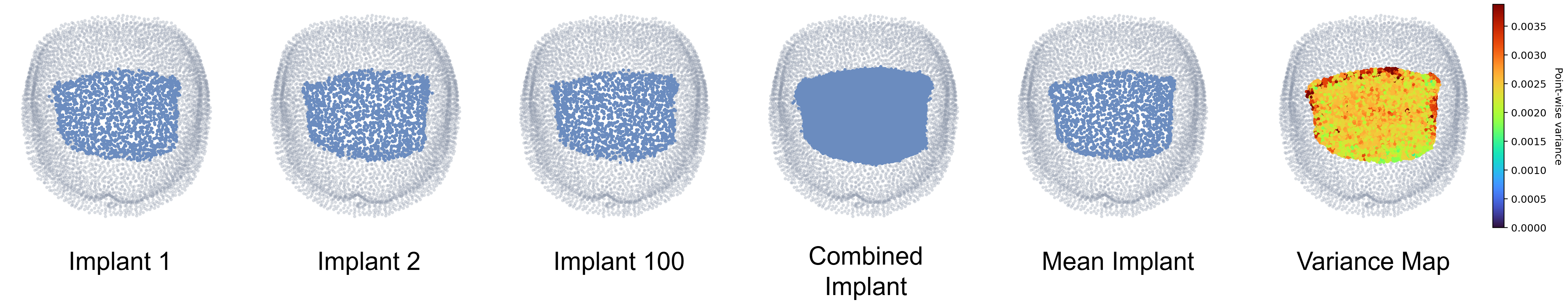}
  \caption{Multi-sample variability for PCFM-PTv3 on a fixed input. We show several generated implant samples, their mean completion, and a variance map. The global implant geometry remains stable across samples, while higher variance appears near the defect boundary, indicating localized uncertainty in ambiguous regions.}
  \label{fig:multisample_variance}
\end{figure}

To visualize this behavior, we generate multiple completions for the same input and compute their mean completion together with an empirical point-wise variance map. The mean implant summarizes the central tendency of the generated candidates, while the variance map highlights regions where the model expresses higher uncertainty. As shown in Figure~\ref{fig:multisample_variance}, the global implant shape remains stable across samples, while variability is mainly localized near the defect boundary. This suggests that PCFM-PTv3 captures uncertainty in boundary placement rather than producing globally unstable reconstructions.

\section{Multiiterative Point Cloud Completion}
\begin{figure}[t]
  \centering
  \includegraphics[width=\linewidth]{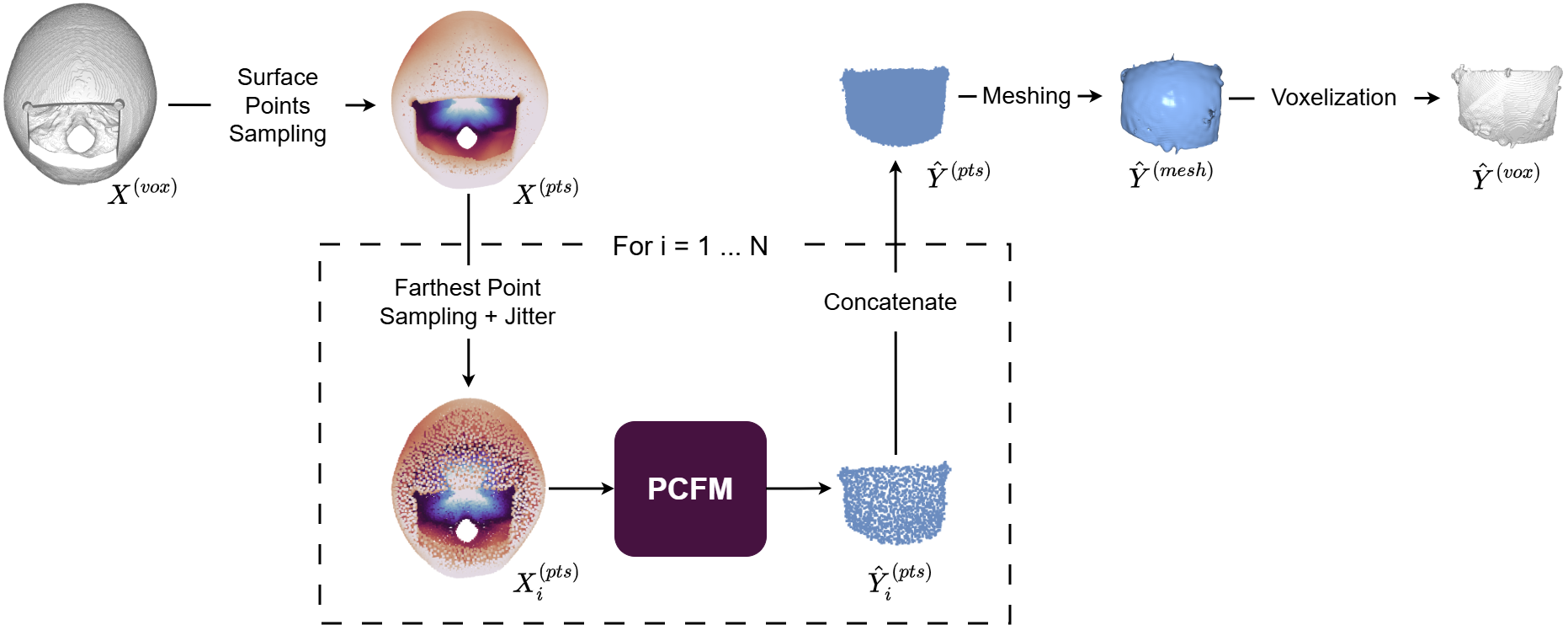}
  \caption{Inference pipeline for the multi-iterative setup. We sample a dense point cloud from the surface and run the model $N$ times on independently resampled (FPS + jitter) inputs. The predicted implant point sets are then aggregated by concatenation to form a denser completion, which is converted to a mesh for downstream fabrication/3D printing and optionally voxelized for volumetric evaluation.}
  \label{fig:multiiter_pipeline}
\end{figure}

In addition to sampling multiple completions from different noise initializations, we also evaluate a multi-iterative inference strategy aimed at improving downstream surface reconstruction. While the previous section analyzes stochastic variability for a fixed input, here we increase the density of the predicted implant point set without modifying the trained model. Starting from a dense surface point cloud, we generate multiple resampled inputs $X_i$ via FPS with jitter and run the completion model independently for $i=1,\dots,N$. The resulting predictions $\hat{Y}_i$ are then aggregated by concatenation to form a denser implant point cloud, which is used for meshing and optionally voxelization for volumetric evaluation. Figure~\ref{fig:multiiter_pipeline} illustrates this pipeline.

\begin{figure}[b]
  \centering
  \includegraphics[width=\linewidth]{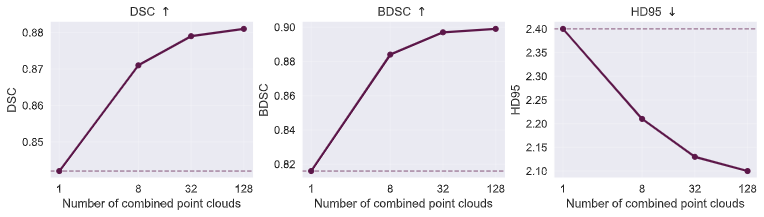}
  \caption{Effect of the multi-iterative aggregation on SkullFix for PCFM--PTv3 (Linear). We run inference on $N\in\{1,8,32,128\}$ independently resampled inputs and concatenate the resulting implant point clouds. Increasing $N$ consistently improves voxel-space metrics (DSC/BDSC $\uparrow$) and reduces HD95 $\downarrow$; the dashed line indicates the single-pass ($N=1$) baseline.}
  \label{fig:multiiterative_skullfix_metrics}
\end{figure}

We report the effect of increasing the number of aggregated runs $N\in\{1,8,32,128\}$ on SkullFix for PCFM--PTv3 (Linear) in Fig.~\ref{fig:multiiterative_skullfix_metrics}. Increasing $N$ consistently improves voxel-space overlap metrics (DSC/BDSC) and reduces HD95, indicating that higher point density can improve the stability of the mesh/voxel conversion step and yield more reliable volumetric evaluation. The gains saturate beyond $N=32$, suggesting that only a moderate number of resamplings is needed in practice. Since this procedure is applied only at inference, it provides a simple way to trade additional runtime for improved surface reconstruction quality without retraining.

\clearpage

\section{Ablation Studies on Affine Paths and Contrastive Term for Flow Matching}
We ablate two key design choices in PCFM: the affine probability path (Linear, VP linear, or Cosine) and the optional contrastive regularizer (CFM). We report Chamfer Distance across model sizes and point settings on SkullFix and SkullBreak in Table~\ref{tab:fm_ablation}, using identical training and evaluation protocols to isolate the effect of each component. Across both datasets and almost all settings, removing the contrastive term (CFM ''$-$'') improves CD, suggesting that $\Delta$FM-style separation is not consistently beneficial for this medical completion task. Among affine schedulers, Cosine and Linear generally outperform VP linear, with the best results obtained at the highest point setting (29492/3276). 

We also observe that increasing the number of points yields the most reliable improvements, whereas scaling model capacity is not strictly monotonic: the Base model is often competitive or best, and the Large model provides limited additional benefit in many configurations. Overall, these ablations motivate our main experimental choice of a non-contrastive PCFM variant with a Linear/Cosine path and sufficiently high point resolution.

\section{Additional Visualizations of Results for the Studied Methods}
\begin{figure}[h]
    \centering
    \includegraphics[width=\textwidth]{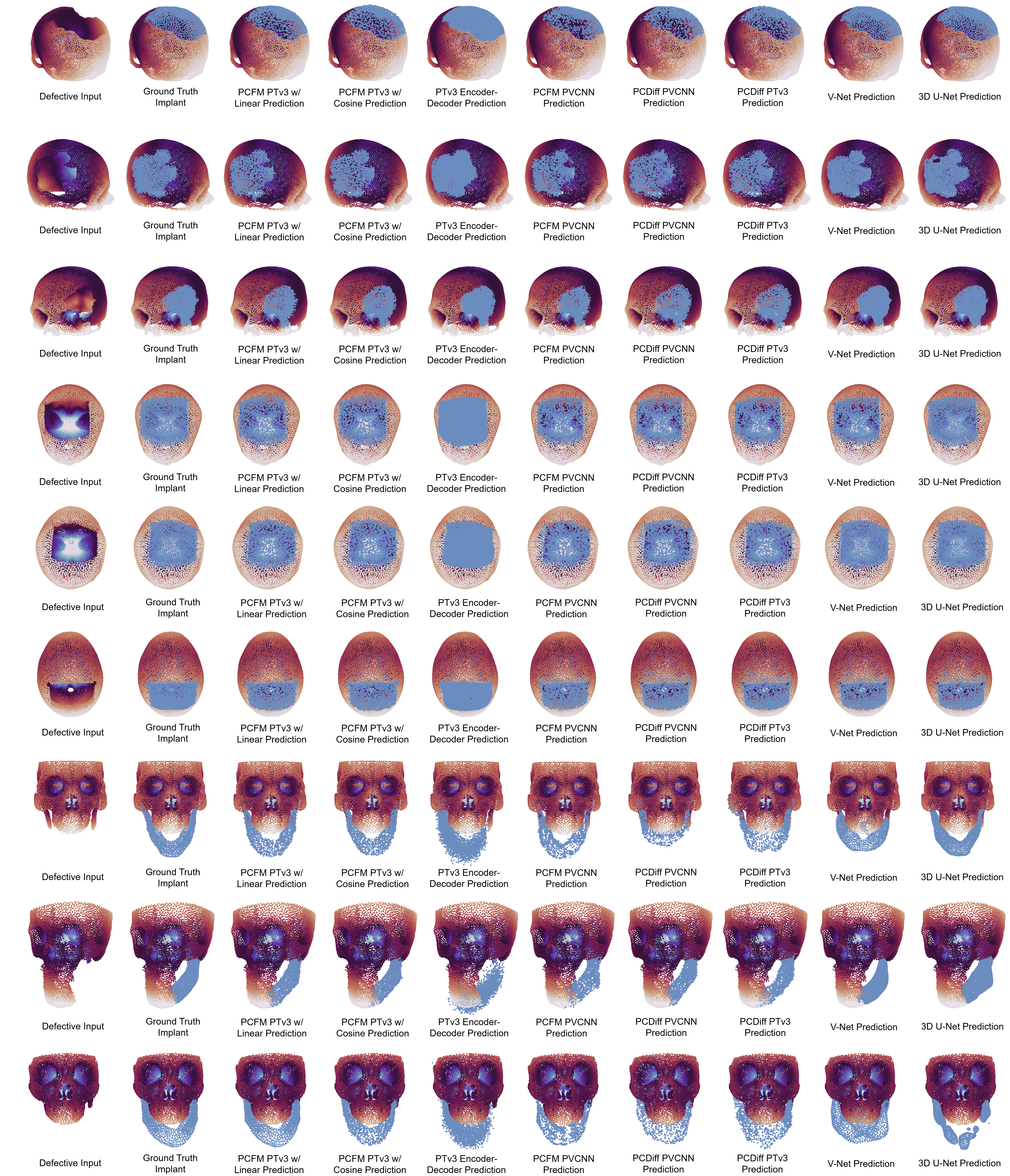}
    \caption{Additional qualitative comparisons across datasets and methods. Rows 1-3: SkullBreak, rows 4-6: SkullFix, rows 7-9: Mandibular Defect. For volumetric baselines, point clouds are obtained by extracting surfaces and sampling points using Poisson surface sampling followed by farthest point sampling (FPS), consistent with the preprocessing used for point-based methods.}
    \label{fig:moreVis}
\end{figure}

We provide additional qualitative results for all compared methods and datasets in Figure~\ref{fig:moreVis}. We show three representative test cases from SkullBreak, SkullFix, and Mandibular Defect (three rows per dataset), illustrating the defective input, the predicted implant, and the corresponding ground truth.

\begin{table}[t]
\centering
\scriptsize
\setlength{\tabcolsep}{4.2pt}
\renewcommand{\arraystretch}{1.05}
\caption{Ablation of affine paths and the contrastive term (CFM) for PCFM across model sizes and point settings. Lower CD is better.}
\label{tab:fm_ablation}
\begin{tabular}{lll c
                S[table-format=1.4]
                S[table-format=1.4]}
\toprule
\textbf{Model} & $\mathbf{N_X/N_Y}$ & \textbf{Affine path} & \textbf{CFM} &
\multicolumn{1}{c}{\textbf{SkullFix CD}$\downarrow$} &
\multicolumn{1}{c}{\textbf{SkullBreak CD}$\downarrow$} \\
\midrule

\multirow{18}{*}{Small}
& \multirow{6}{*}{1844/204}
& \multirow{2}{*}{Linear} & + & 0.2640 & 0.3170 \\
&                        &        & - & 0.2450 & 0.3070 \\
&                        & \multirow{2}{*}{VP linear} & + & 0.2820 & 0.3790 \\
&                        &        & - & 0.2690 & 0.3440 \\
&                        & \multirow{2}{*}{Cosine} & + & 0.2590 & 0.3830 \\
&                        &        & - & 0.2590 & 0.3660 \\
\cmidrule(lr){2-6}
& \multirow{6}{*}{14746/1638}
& \multirow{2}{*}{Linear} & + & 0.2260 & 0.1970 \\
&                         &        & - & 0.1290 & 0.1300 \\
&                         & \multirow{2}{*}{VP linear} & + & 0.2110 & 0.2010 \\
&                         &        & - & 0.1570 & 0.1450 \\
&                         & \multirow{2}{*}{Cosine} & + & 0.1810 & 0.1700 \\
&                         &        & - & 0.1390 & 0.1330 \\
\cmidrule(lr){2-6}
& \multirow{6}{*}{29492/3276}
& \multirow{2}{*}{Linear} & + & 0.2060 & 0.1910 \\
&                         &        & - & 0.1030 & 0.1040 \\
&                         & \multirow{2}{*}{VP linear} & + & 0.1910 & 0.1750 \\
&                         &        & - & 0.1290 & 0.1180 \\
&                         & \multirow{2}{*}{Cosine} & + & 0.1640 & 0.1540 \\
&                         &        & - & 0.1060 & 0.1060 \\
\midrule

\multirow{18}{*}{Base}
& \multirow{6}{*}{1844/204}
& \multirow{2}{*}{Linear} & + & 0.2630 & 0.3280 \\
&                        &        & - & 0.2440 & 0.3050 \\
&                        & \multirow{2}{*}{VP linear} & + & 0.2720 & 0.3310 \\
&                        &        & - & 0.2650 & 0.3050 \\
&                        & \multirow{2}{*}{Cosine} & + & 0.2520 & 0.3180 \\
&                        &        & - & 0.2590 & 0.3090 \\
\cmidrule(lr){2-6}
& \multirow{6}{*}{14746/1638}
& \multirow{2}{*}{Linear} & + & 0.2120 & 0.1870 \\
&                         &        & - & 0.1130 & 0.1200 \\
&                         & \multirow{2}{*}{VP linear} & + & 0.1680 & 0.1680 \\
&                         &        & - & 0.1370 & 0.1370 \\
&                         & \multirow{2}{*}{Cosine} & + & 0.1600 & 0.1490 \\
&                         &        & - & 0.1180 & 0.1220 \\
\cmidrule(lr){2-6}
& \multirow{6}{*}{29492/3276}
& \multirow{2}{*}{Linear} & + & 0.2160 & 0.1900 \\
&                         &        & - & \textbf{0.0939} & 0.0997 \\
&                         & \multirow{2}{*}{VP linear} & + & 0.1520 & 0.1570 \\
&                         &        & - & 0.1050 & 0.1090 \\
&                         & \multirow{2}{*}{Cosine} & + & 0.1580 & 0.1460 \\
&                         &        & - & 0.1020 & \textbf{0.0988} \\
\midrule

\multirow{18}{*}{Large}
& \multirow{6}{*}{1844/204}
& \multirow{2}{*}{Linear} & + & 0.2540 & 0.2750 \\
&                        &        & - & 0.2370 & 0.2900 \\
&                        & \multirow{2}{*}{VP linear} & + & 0.2800 & 0.3070 \\
&                        &        & - & 0.2660 & 0.2810 \\
&                        & \multirow{2}{*}{Cosine} & + & 0.2410 & 0.2800 \\
&                        &        & - & 0.2460 & 0.2900 \\
\cmidrule(lr){2-6}
& \multirow{6}{*}{14746/1638}
& \multirow{2}{*}{Linear} & + & 0.2170 & 0.1870 \\
&                         &        & - & 0.1110 & 0.1130 \\
&                         & \multirow{2}{*}{VP linear} & + & 0.1730 & 0.1670 \\
&                         &        & - & 0.1270 & 0.1300 \\
&                         & \multirow{2}{*}{Cosine} & + & 0.1540 & 0.1420 \\
&                         &        & - & 0.1070 & 0.1140 \\
\cmidrule(lr){2-6}
& \multirow{6}{*}{29492/3276}
& \multirow{2}{*}{Linear} & + & 0.2590 & 0.2210 \\
&                         &        & - & 0.0997 & 0.1100 \\
&                         & \multirow{2}{*}{VP linear} & + & 0.1760 & 0.1620 \\
&                         &        & - & 0.1240 & 0.1300 \\
&                         & \multirow{2}{*}{Cosine} & + & 0.1700 & 0.1550 \\
&                         &        & - & 0.0974 & 0.1010 \\
\bottomrule
\end{tabular}
\end{table}

\clearpage

\end{document}